\documentclass[letterpaper, 10 pt, conference]{ieeeconf}  % Comment this line out if you need a4paper

\IEEEoverridecommandlockouts                              % This command is only needed if 
\usepackage{graphics} % for pdf, bitmapped graphics files
\usepackage{epsfig} % for postscript graphics files
\usepackage{amsmath} % assumes amsmath package installed
\usepackage{amsfonts} % assumes amsmath package installed
\usepackage{amssymb}  % assumes amsmath package installed
\usepackage{mathtools}
\usepackage{bbm}
\usepackage[hidelinks]{hyperref}
\usepackage{siunitx}
\usepackage{import}
\usepackage{booktabs}
\usepackage{placeins}
\usepackage{tensor}
\usepackage{tikz}
\usepackage{bigdelim}
\usepackage{array}
\usepackage{gensymb}
\usepackage{makecell}
\usepackage{threeparttable}
\usepackage{colortbl}
\usepackage{boldline}
\IEEEtriggeratref{7}

\newcolumntype{L}[1]{>{\raggedright\let\newline\\\arraybackslash\hspace{0pt}}m{#1}}
\newcolumntype{C}[1]{>{\centering\let\newline\\\arraybackslash\hspace{0pt}}m{#1}}
\newcolumntype{R}[1]{>{\raggedleft\let\newline\\\arraybackslash\hspace{0pt}}m{#1}}

\title{\LARGE \bf SDF-based RGB-D Camera Tracking in Neural Scene Representations}

\author{Leonard Bruns$^{1}$ and Fereidoon Zangeneh$^{1,2}$ and Patric Jensfelt$^{1}$% <-this % stops a space
\thanks{$^{1}$The authors are with the Division of Robotics, Perception and Learning (RPL),
KTH Royal Institute of Technology, Stockholm, Sweden {\tt\small\{leonardb,fzk,patric\}@kth.se}.}
\thanks{$^{2}$ Fereidoon Zangeneh is also with Univrses AB.}%
}

\begin{document}

\maketitle
\thispagestyle{empty}
\pagestyle{empty}

%%%%%%%%%%%%%%%%%%%%%%%%%%%%%%%%%%%%%%%%%%%%%%%%%%%%%%%%%%%%%%%%%%%%%%%%%%%%%%%%
\begin{abstract}

We consider the problem of tracking the 6D pose of a moving RGB-D camera in a neural scene representation. Different such representations have recently emerged, and we investigate the suitability of them for the task of camera tracking. In particular, we propose to track an RGB-D camera using a signed distance field-based representation and show that compared to density-based representations, tracking can be sped up, which enables more robust and accurate pose estimates when computation time is limited.
\vspace*{0.3cm}
\end{abstract}

{
\setlength{\parskip}{3pt}
\section{Introduction}
% Camera tracking is a crucial part of an integrated visual SLAM system, which allows an agent to localize itself in an environment. Camera tracking has been the subject of extensive research in the past three decades, with solutions relying on techniques such as direct alignment of images or tracking of sparse salient feature points that constitute the map.

Recently, neural scene representations have been shown to possess promising characteristics for creating dense reconstructions of environments \cite{mildenhall2020nerf, wang2021neus, azinovic2022neural}. The continuous map stored in the weights of these coordinate-based networks can densely represent environments through quantities such as radiance fields \cite{mildenhall2020nerf}, occupancy probability \cite{mescheder2019occupancy, zhu2021nice}, and signed-distance fields (SDFs) \cite{park2019deepsdf, wang2021neus, yariv2021volume, ortiz2022isdf}.

To map an environment with a neural scene representation using a camera with unknown trajectory, one needs to perform camera tracking \cite{wang2021nerf, sucar2021imap}. Camera tracking can be done by direct comparison of observations and renders of the mapped scene. Rendering of views is typically done via volume rendering, which involves densely querying the viewing frustum. This requires many samples and is, therefore, time consuming.

In this paper, we investigate whether recently proposed SDF-based neural scene representations \cite{wang2021neus, yariv2021volume} can be used for more efficient tracking when paired with RGB-D cameras compared to volume rendering-based tracking, such as in iMAP \cite{sucar2021imap}. We propose a novel tracking scheme that estimates the camera pose by directly querying the observed surface points and minimizing the returned distances. This obviates the need for volume rendering, increasing the time budget that instead can be used for incorporating more of the observations. % We show that this results in more robust tracking.

\subsection{Problem Definition}
Given an initial camera pose $\tensor*[^{\mathrm{w}}_0]{\mathbf{T}}{_{\mathrm{c}}}$ and a stream of RGB-D images $(\mathbf{I}_i\in\mathbb{R}^{H\times W \times 3}, \mathbf{D}_i\in\mathbb{R}^{H\times W}), i=1,...,M$ we want to find estimates $\tensor*[^{\mathrm{w}}_i]{\widetilde{\mathbf{T}}}{_{\mathrm{c}}}$ of the true camera poses $\tensor*[^{\mathrm{w}}_i]{\mathbf{T}}{_{\mathrm{c}}}$. We assume a known static environment that has previously been encoded in a neural scene representation.
}

\newpage

\begin{figure}[t!]
    \centering
    \includegraphics[width=0.7\linewidth]{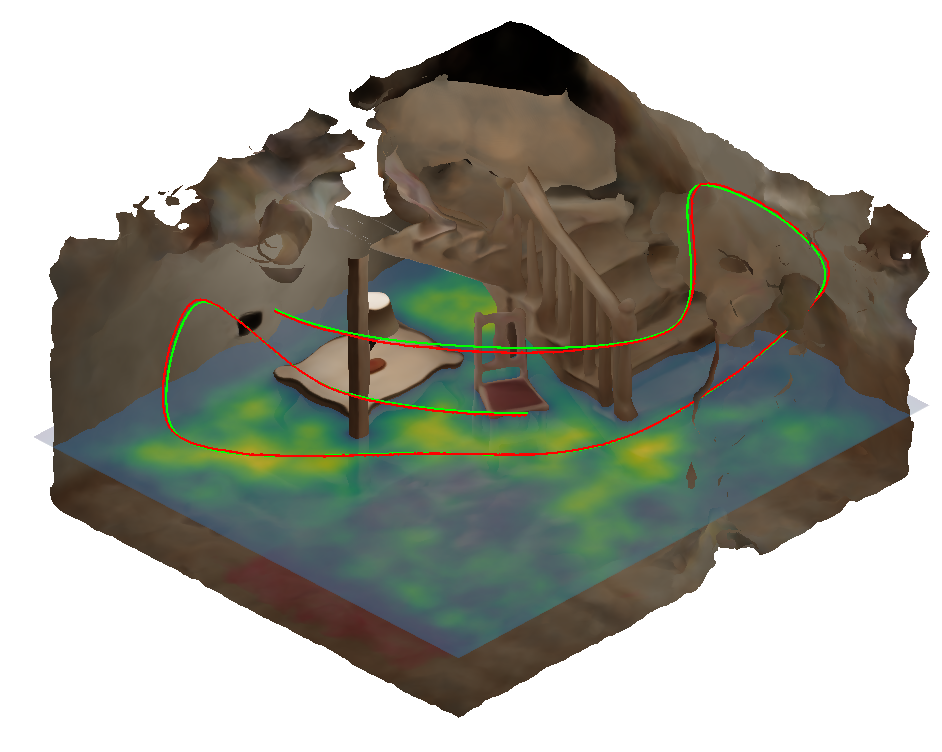}
    \caption{True (\protect\tikz [baseline=-0.5ex]{\protect\draw[green,line width=1pt] (0,0) -- ++ (0.4,0);}) and tracked (\protect\tikz [baseline=-0.5ex]{\protect\draw[red,line width=1pt] (0,0) -- ++ (0.4,0);}) camera trajectory and signed distance field used for tracking. The mesh was extracted from the zero-isosurface of the signed distance field.}\label{fig:cooltrajectory}
\end{figure}

\section{Method}

\subsection{Scene Representations}
\begin{table*}
    \centering
    \caption{Root-mean-square average trajectory error (in \SI{}{m}) on the neural RGB-D dataset by \cite{azinovic2022neural} ($\color{red}>\SI{0.2}{m}$, \textbf{better}).}\label{tab:ate}
    \scriptsize
    \begin{tabular}{@{}llR{1cm}R{1cm}R{1cm}R{1cm}R{1cm}R{1cm}R{1cm}R{1cm}R{1cm}@{}}
        \toprule
         & & Breakfast & Compl. Kitchen & Green Room & Grey-white & Kitchen & Morning Apart. & Staircase & Thin Geo. & Whiteroom\\
        \midrule
        \multirow{2}{*}{$r=\SI{10}{Hz}$}& iMAP with VR & $\color{red} 1.575$ & $\color{red} 2.171$ & $\color{red} 0.702$ & $\mathbf{0.156}$ & $\color{red} 2.196$ & $\mathbf{0.107}$ & $\color{red} 3.582$ & $\color{red} 0.312$ & $\color{red} 2.744$ \\
        & NeuS with SDF & $\mathbf{0.096}$ & $\mathbf{0.118}$ & $\color{red} \mathbf{0.215}$ & $\color{red} 0.255$ & $\mathbf{0.110}$ & $0.114$ & $\mathbf{0.052}$ & $\mathbf{0.031}$ & $\mathbf{0.067}$ \\
        \midrule
        \multirow{2}{*}{$r=\SI{5}{Hz}$}& iMAP with VR & $0.082$ & $0.105$ & $\mathbf{0.048}$ & $\mathbf{0.032}$ & $\color{red} 1.313$ & $\mathbf{0.027}$ & $\color{red} 3.868$ & $0.035$ & $0.080$ \\
        & NeuS with SDF & $\mathbf{0.043}$ & $\mathbf{0.049}$ & $0.100$ & $0.072$ & $\mathbf{0.044}$ & $0.073$ & $\mathbf{0.045}$ & $\mathbf{0.017}$ & $\mathbf{0.022}$ \\
        \midrule
        \multirow{2}{*}{$r=\SI{2}{Hz}$}& iMAP with VR & $\mathbf{0.021}$ & $\mathbf{0.020}$ & $\mathbf{0.015}$ & 0.077 & $\mathbf{0.077}$ & $\mathbf{0.007}$ & $\color{red}3.987$ & 
        $\mathbf{0.012}$ & 0.022 \\
        & NeuS with SDF & 0.038 & 0.038 & 0.067 & $\mathbf{0.043}$ & $\color{red}1.148$ & 0.067 & $\mathbf{0.044}$ & 0.014 & $\mathbf{0.015}$ \\
        % 10\degree,\SI{2}{cm} & 0.331 & \textbf{0.535} & 0.331 & & 0.013 & 0.2 & \textbf{0.307} \\
        % 5\degree,\SI{1}{cm} & 0.073 & \textbf{0.205} & 0.069 & & 0.000 & 0.013 & \textbf{0.080} \\
        % 10\degree,\SI{2}{cm},0.6 & 0.031 & \textbf{0.471} & 0.215 & & 0.000 & \textbf{0.173} & \textbf{0.173} \\
        % 5\degree,\SI{1}{cm},0.8 & 0.000 & \textbf{0.170} & 0.050 & & 0.000 & 0.013 & \textbf{0.053} \\
        \bottomrule
    \end{tabular}
\end{table*}

\begin{table}
    \centering
    \caption{Average number of optimization steps under varying time constraints and numbers of pixel samples $n$.}\label{tab:runtime}
    \scriptsize
    \begin{tabular}{@{}lcR{1cm}R{1cm}R{1cm}@{}}
        \toprule
        & $n$ & \SI{50}{ms} & \SI{100}{ms} & \SI{500}{ms} \\
        \midrule
        \multirow{3}{*}{VR}& $128$ & 3.2 & 6.6 & 36.2\\
        & $\mathbf{512}$ & 3.2 & 6.8 & 36.1\\
        & $1024$ & 2.2 & 5.0 & 27.0\\
        \midrule
        \multirow{3}{*}{SDF} & $2048$ & 7 & 15 & 77 \\
         & $\mathbf{4096}$ & 7.1 & 15.3 & 79.3 \\
         & $16384$ & 5.6 & 12.7 & 65.8 \\
        \bottomrule
    \end{tabular}
\end{table}

% TODO any nice way of vertically centering the *and*
% this seems to have some solutions, albeit ugly:
% https://tex.stackexchange.com/questions/477547/two-functions-in-the-same-line
We consider two neural scene representations: iMAP \cite{sucar2021imap} and NeuS \cite{wang2021neus}. We denote their respective networks by
\begin{equation}
    \begin{alignedat}{2}
        f_\mathrm{iMAP}: \mathbb{R}^{3} &\to \mathbb{R}^3\times \mathbb{R} \quad \text{and} \quad f_\mathrm{NeuS}:\, & \mathbb{R}^{3} &\to \mathbb{R}^3\times \mathbb{R}\\
        \mathbf{x}&\mapsto (\mathbf{c}, \sigma) & \mathbf{x} &\mapsto (\mathbf{c}, d),
    \end{alignedat}
\end{equation}
where $\mathbf{c}$ denotes color, $\sigma$ density, and $d$ the signed distance to the closest surface.

\subsection{Camera Tracking}
\subsubsection{Density-based Rendering}
In density-based representations (such as iMAP or NeRF), the camera pose can be optimized by rendering the color and depth image at the previous estimate and formulating a loss by comparing the rendered and observed color and depth images. This has previously been demonstrated by iNeRF \cite{yen2021inerf} for RGB data only and with RGB-D data by iMAP \cite{sucar2021imap}. Due to the expensive nature of volume rendering, both of these methods only render a small subset of pixels to reduce the computation time. We use volume rendering as described in iMAP as our baseline.

\subsubsection{SDF-based Optimization}
Under ideal conditions, all the observed depth points originate from surfaces in the environment. Therefore, these points should fall onto zero crossings of the SDF. We propose to exploit this synergy between RGB-D cameras and SDFs by directly querying the scene representation at the points from the RGB-D point cloud after transforming it into the world frame using the previous pose estimate. Instead of defining a loss based on the render differences, we formulate the loss based on the query points' colors and signed distances.

Specifically, we sample $n$ points from the current RGB-D image $(\mathbf{I}_i, \mathbf{D}_i)$ and compute the colored point set $\tensor*[^{\mathrm{c}}]{\mathcal{P}}{_i}=\{(\tensor*[^{\mathrm{c}}]{\mathbf{p}}{_k}\in \mathbb{R}^3,\tensor*{\mathbf{c}}{_k}\in \mathbb{R}^3)| k=1,...,n\}$ in the camera frame. We then optimize the loss
\begin{equation}
    l = \lambda_\mathrm{SDF} \frac{1}{n}\sum_{k=1}^n |\tilde{d}_k| + \lambda_\mathrm{color} \frac{1}{3n} \sum_{k=1}^n \lVert\tilde{\mathbf{c}}_k - \mathbf{c}_k \rVert_1,
\end{equation}
where $(\tilde{\mathbf{c}}_k, \tilde{d}_k )=f_\mathrm{NeuS}(\tensor*[^{\mathrm{w}}_i]{\widetilde{\mathbf{T}}}{_{\mathrm{c}}}\tensor*[^{\mathrm{c}}]{\mathbf{p}}{_k})$, and $\lambda_\mathrm{SDF}$ and $\lambda_\mathrm{color}$ are fixed hyperparameters. Similarly to iMAP, we sample a new set of $n$ points for every optimization iteration.

Note that the same optimization is not applicable to density-based representations, which typically converge to very sharp boundaries without smooth spatial gradients that could guide the optimization (see Fig.~\ref{fig:cooltrajectory} for an example of a learned SDF). Furthermore, the isosurface on which the depth points lie is undefined. Similarly, occupancy fields \cite{mescheder2019occupancy}, despite having a well-defined isosurface at 0.5, converge to very sharp transitions under ideal training conditions.

% We propose to optimize the current at time $i$ camera pose $\tensor[^{\mathrm{w}}]{\widetilde{\mathbf{T}}}{_{\mathrm{c}}}$

\section{Experiments}

\subsubsection{Implementation Details} We use the same network architecture as iMAP for both methods. NeuS contains a single additional trainable parameter for the standard deviation of the s-density \cite{wang2021neus}. Both methods are implemented in PyTorch. We parametrize the camera pose as a position $\tensor*[^{\mathrm{w}}_i]{\tilde{\mathbf{t}}}{_{\mathrm{c}}}\in \mathbb{R}^3$ and unit quaternion $\tensor*[^{\mathrm{w}}_i]{\tilde{\mathbf{q}}}{_{\mathrm{c}}}\in \mathbb{H}_1$ (we renormalize after every optimization step). We use Adam optimizer \cite{kingma2014adam} with learning rates $5\times 10^{-4}$ and $1\times 10^{-3}$ for position and orientation, respectively.

\subsubsection{Evaluation Protocol}
We report the root-mean-square of the absolute trajectory error (ATE) 
\begin{equation}
    ATE = \sqrt{\frac{1}{M}\sum_{i=1}^{M}\lVert\tensor*[^{\mathrm{w}}_i]{\tilde{\mathbf{t}}}{_{\mathrm{c}}} - \tensor*[^{\mathrm{w}}_i]{\mathbf{t}}{_{\mathrm{c}}}\rVert_2^2 },
\end{equation}
where $\tensor*[^{\mathrm{w}}_i]{\tilde{\mathbf{t}}}{_{\mathrm{c}}}$ and $\tensor*[^{\mathrm{w}}_i]{\mathbf{t}}{_{\mathrm{c}}}$ denote the translation part of $\tensor*[^{\mathrm{w}}_i]{\widetilde{\mathbf{T}}}{_{\mathrm{c}}}$ and $\tensor*[^{\mathrm{w}}_i]{\mathbf{T}}{_{\mathrm{c}}}$, respectively.
We process every frame in the sequence with a fixed tracking rate $r$ (i.e., we do not force a certain playback frame rate or drop frames due to too slow tracking). For each frame, we estimate the average time per optimization iteration and continue to the next frame if the remaining time budget is not sufficient for another iteration.

\subsubsection{Results}
We report results on the sequences of the dataset by \cite{azinovic2022neural} in Table \ref{tab:ate}. \emph{NeuS with SDF} refers to our proposed SDF-based tracking and \emph{iMAP with VR} refers to iMAP with volume rendering as described in \cite{sucar2021imap}.

We can see that our proposed SDF-based loss fails less frequently when tracking at faster frame rates. The results for $r=\SI{2}{Hz}$ indicate that volume rendering achieves lower ATE when sufficient optimization time is available. We hypothesize that tracking using volume rendering might be more accurate here, since the underlying representation was trained via volume rendering as well. By contrast, our SDF-based tracking loss differs significantly from the training time loss.

In Table \ref{tab:runtime} we further show the number of iterations for different time budgets and number of samples $n$. Note that because our SDF-based tracking does not rely on expensive volume rendering, we can incorporate more of the available sensor data into each optimization step.

\section{Conclusion}
Our experiments confirm that an SDF-based representation can be used to more efficiently track an RGB-D camera inside a neural scene representation. This comes at the cost of a more involved mapping task, which in our case involved the eikonal term \cite{gropp2020implicit}, which requires the computation of second-order gradients and therefore slows down training roughly by a factor of two. In the future we want to investigate whether the SDF loss can similarly be used to speed up mapping by giving direct supervision to the isosurface in combination with volume rendering.

\section*{Acknowledgment}
This work was partially supported by the Wallenberg AI, Autonomous Systems and Software Program (WASP) funded by the Knut and Alice Wallenberg Foundation. 

% \nocite{*}
\clearpage

\bibliographystyle{IEEEtran}
\bibliography{IEEEabrv,bib}

\end{document}